\documentclass[letterpaper, 10 pt, conference]{ieeeconf}
\pdfoutput=1
\IEEEoverridecommandlockouts
\usepackage{cite}
\usepackage{amsmath,amssymb,amsfonts}
\usepackage{algorithmic}
\usepackage{graphicx}
\graphicspath{{../figures/}}
\usepackage{textcomp}
\usepackage{xcolor}
\usepackage{mathptmx}
\usepackage{array}
\usepackage{dblfloatfix}

\def\BibTeX{{\rm B\kern-.05em{\sc i\kern-.025em b}\kern-.08em
    T\kern-.1667em\lower.7ex\hbox{E}\kern-.125emX}}

\usepackage{fancyhdr}
\begin{document}

\title{Unsupervised Quality Prediction for Improved Single-Frame and Weighted Sequential Visual Place Recognition 
\thanks{This research is partially supported by an ARC Laureate Fellowship FL210100156 to MM, the QUT Centre for Robotics, the Centre for Advanced Defence Research in Robotics and Autonomous Systems, and received funding from the Australian Government via grant AUSMURIB000001 associated with ONR MURI grant N00014-19-1-2571. The work of H.Carson was supported in part by an Australian Postgraduate Award.}
\thanks{The authors are with the QUT Centre for Robotics, School of Electrical Engineering and Robotics at the Queensland University of Technology, Brisbane, Australia (e-mail: h.carson@hdr.qut.edu.au, j2.ford@qut.edu.au, michael.milford@qut.edu.au).}
}
\author{Helen Carson, Jason J. Ford, Michael Milford}

\maketitle
\thispagestyle{fancy}
\pagestyle{plain}

\begin{abstract}

While substantial progress has been made in the absolute performance of localization and Visual Place Recognition (VPR) techniques, it is becoming increasingly clear from translating these systems into applications that other capabilities like integrity and predictability are just as important, especially for safety- or operationally-critical autonomous systems. In this research we present a new, training-free approach to predicting the likely quality of localization estimates, and a novel method for using these predictions to bias a sequence-matching process to produce additional performance gains beyond that of a naive sequence matching approach. Our combined system is lightweight, runs in real-time and is agnostic to the underlying VPR technique. On extensive experiments across four datasets and three VPR techniques, we demonstrate our system improves precision performance, especially at the high-precision/low-recall operating point. We also present ablation and analysis identifying the performance contributions of the prediction and weighted sequence matching components in isolation, and the relationship between the quality of the prediction system and the benefits of the weighted sequential matcher.
\end{abstract}

\section{Introduction}

Visual Place Recognition (VPR) is the task of estimating position using cameras, by matching an image of the current scene to a database of geo-tagged reference images taken along a previous traverse of the route \cite{Lowry2016}. Substantial progress has been made in this research area in recent years~\cite{Yin22Survey, Yasuda20,Zhang2020}. While absolute performance has often been the target, when adapting these systems to operate on deployed robots and autonomous systems, other capabilities around localization integrity become just as, if not more, important.

While substantial progress has been made in other domains including aerospace on topics like system integrity \cite{DO229}, this field is relatively underinvestigated in robotics. Initial pilot work in this area has demonstrated some promise, but has had substantial practical limitations including the need for extensive, multi-traverse training data with ground truth correspondences from the target domain. In this paper, we present a new VPR system that provides high performance whilst overcoming some of the practical limitations of previous approaches.

\begin{figure}[!t]
\centering
\includegraphics[width=\columnwidth]{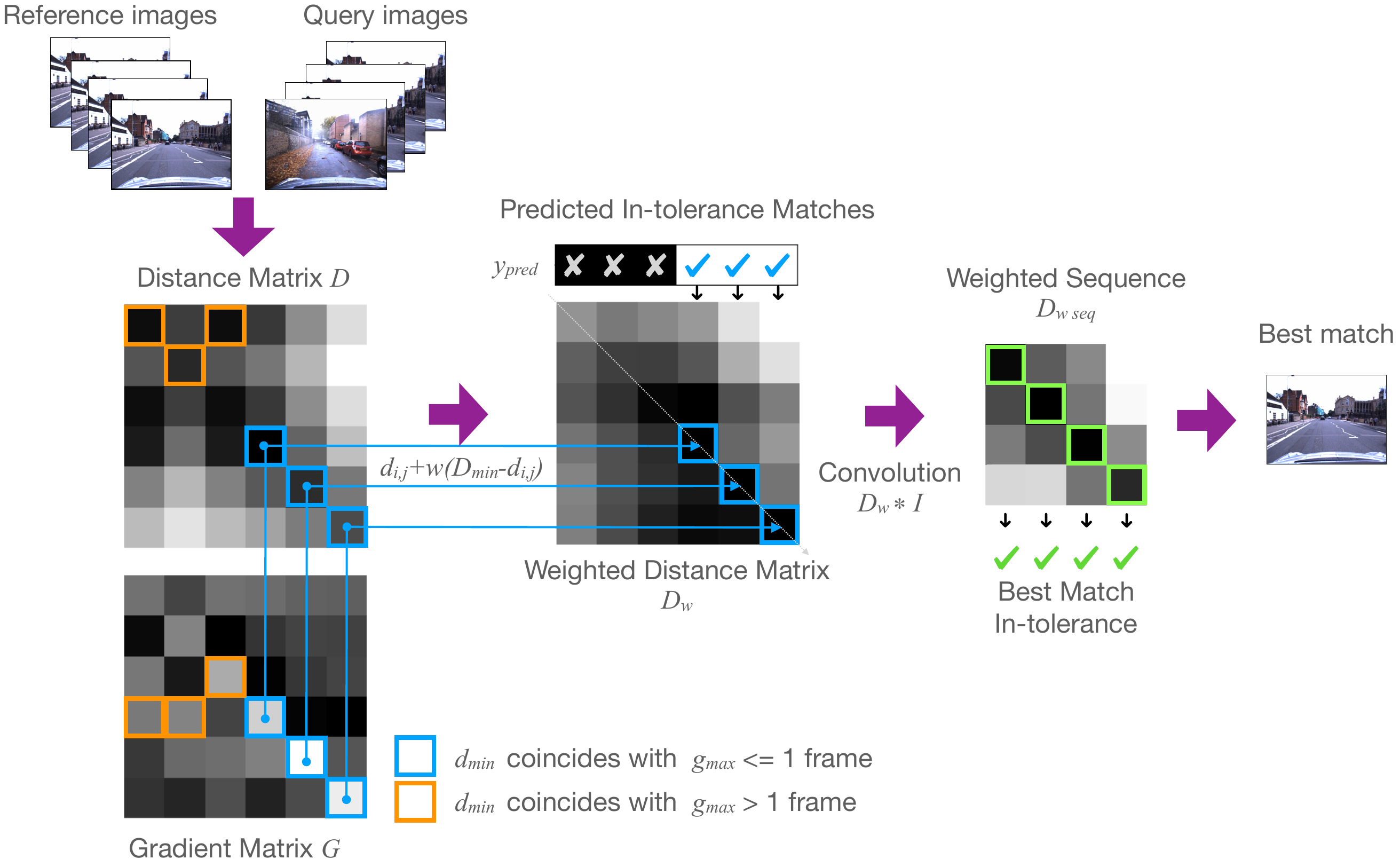}
\caption{Visual representation of our combined approach. For each query, matches are predicted to be in-tolerance when the distance matrix minimum and gradient matrix maximum coincide within one frame. These consensus matches are weighted heavily before applying sequence matching, resulting in a higher proportion of sequence matches that are in-tolerance.}
\label{fig:cartoon}
\end{figure}

We make the following contributions:
\begin{enumerate}
    \item We present a novel, unsupervised method that enables out-of-tolerance matches to be identified and discarded during a post-processing step on various VPR techniques, using a consensus-based approach.
    \item We use our localization quality prediction technique to anchor sequence-based matching around identifiably good localization points, further improving performance
    (Fig. \ref{fig:cartoon}).
\end{enumerate}

To support our contribution claims, we demonstrate increased average precision over 15 combinations of benchmark datasets and VPR techniques, covering viewpoint variation, seasonal changes, and urban driving scenarios, using both simple and state-of-the-art VPR techniques.

We further provide analysis isolating the performance contributions of both the prediction system and the weighted sequence matching technique, as well as of the relationship between the performance of the prediction system and the subsequent effect on performance of the weighted sequence matcher.

The paper proceeds as follows. We provide background in Section \ref{sec:background} and describe our approach in Section \ref{sec:approach}. Our experimental setup is outlined in Section \ref{sec:method}, and results and discussion in Sections \ref{sec:results} and \ref{sec:discussion}.

\section{Background}\label{sec:background}

In this section, we further define the problem, provide an overview of VPR techniques and related works, and outline the differences between our methods and similar post-processing techniques.

\subsection{VPR Techniques}

VPR techniques themselves take many forms, including direct image comparison, hand-crafted techniques, machine-learned feature extraction, hierarchical methods, and semantic segmentation \cite{Yin22Survey}. All involve a method of describing images in the form of extracted features prior to comparison, ranging from simple downsampling and patch normalization \cite{SeqSLAM}, to complex techniques such at SIFT \cite{SIFT99}, SURF \cite{SURF08},ORB \cite{ORB11}, FAB-MAP \cite{Cummins2010FABMAPAP}, Convolutional Neural Network (CNN) architectures \cite{Chen17} such as NetVLAD \cite{NetVLAD18}; and hierarchical methods such as PatchNetVLAD \cite{PatchNetVLAD}. A localization match is then formed by finding the minimum distance between features in the current scene and all reference images. This match however can be incorrect due to visual aliasing, changes in the environment, dynamic agents, lighting differences and
shadows, and many other issues that cause images taken at the same place to look different, and different places to look the same \cite{Lowry2016}.

Applying post-processing techniques to initial VPR estimates can result in large performance improvements \cite{2021Schubert}. These can include fusion with additional sensor data \cite{Tsintotas2021}; use of visual odometry to reject out-of-tolerance estimates \cite{kanji2016multi}; temporal window-based filtering \cite{Bruce2017}; and Random Sample Consensus (RANSAC) to identify outliers once an error tolerance threshold is tuned \cite{RANSAC}. Sequential filtering can also be used to exploit the rich spatio-temporal information provided as a localization system moves along a route, rather than treating each image independently \cite{SeqSLAM}, including learned descriptors based on contrastive learning methods such as SeqNet \cite{Garg21_SeqNet} and SeqMatchNet \cite{garg2021seqmatchnet}. Extended Kalman Filters (EKFs) \cite{EKF22} and particle filters \cite{particlefilters18} can also be used to provide continually updated localization estimates, based on motion state-space estimation. 

\subsection{Localization Integrity}
We relate our work to the concept of integrity, which aims to provide a real-time determination of whether a current localization estimate is within an acceptable tolerance from ground-truth, or not \cite{Hage21}.
As the overall performance of camera-based localization systems becomes higher, and these systems are able to perform reliably in more challenging environments, a system's ability to self-identify when its performance is poor is becoming increasingly important, especially in safety- or operationally-critical applications like autonomous vehicles \cite{Zhu22}.

In our previous work \cite{Carson22}, we proposed a supervised learning method for predicting which matches along a route are in-tolerance, based on extracting features from the distance matrix. 
However, this supervised technique had substantial practical disadvantages: it required two traverses along a route through a representative environment, along with ground-truth data for each reference and query frame, to enable the prediction method to learn which points were likely out-of-tolerance.

Here we provide two complementary techniques: an unsupervised method for predicting in-tolerance matches, and a method of weighting sequences prior to matching based on these predictions to improve overall precision. In contrast to our previous work, this technique requires no calibration or training data.
Our unsupervised method is effectively a consensus-based filter used to ``check'' the distance-based localization estimate from the VPR technique against a secondary and independent gradient-based estimate. Both of these techniques are suited to applications where high precision and low false positive rates are more important than recall - that is, where it is better to reject good points as bad, rather than the alternative: accepting out-of-tolerance estimates as good.

In the following sections we outline our approach and experimental method, and compare the performance of our proposed techniques to baseline VPR techniques. 

\section{Approach}\label{sec:approach}

In this section we detail our approach to our two contributions, including the formulation of a distance matrix, our new gradient matrix, consensus process, and the exploitation of this process to create an improved weighted sequence matching approach.

\subsection{Unsupervised prediction}

The cornerstone of our unsupervised method is based on the observation that the absolute average gradient of the similarity matrix is typically maximised around the best match. This is most evident when creating a similarity matrix of one set of images against itself: all exact matches produce an instantaneous distance of zero along the diagonal, corresponding with a sharp instantaneous rate of change in the distance vector at each matching point. Within the matrix, similar places may exhibit relatively low distances between each other, but intuitively, the absolute average gradient change should be highest around the closest match when the distance is closest to zero. 

Our goal is to identify correct (in-tolerance) points with high precision, such that inaccurate points are discarded and only accurate points remain. Based on our observations, we form the hypothesis that points along a route where the minimum distance and maximum rate-of-change (gradient) of the distance vector coincide are more likely to be exact or very close matches, rather than where these two phenomenon occur at vastly different positions along the route. 

\begin{figure}[!t]
\centering
\includegraphics[width=\columnwidth]{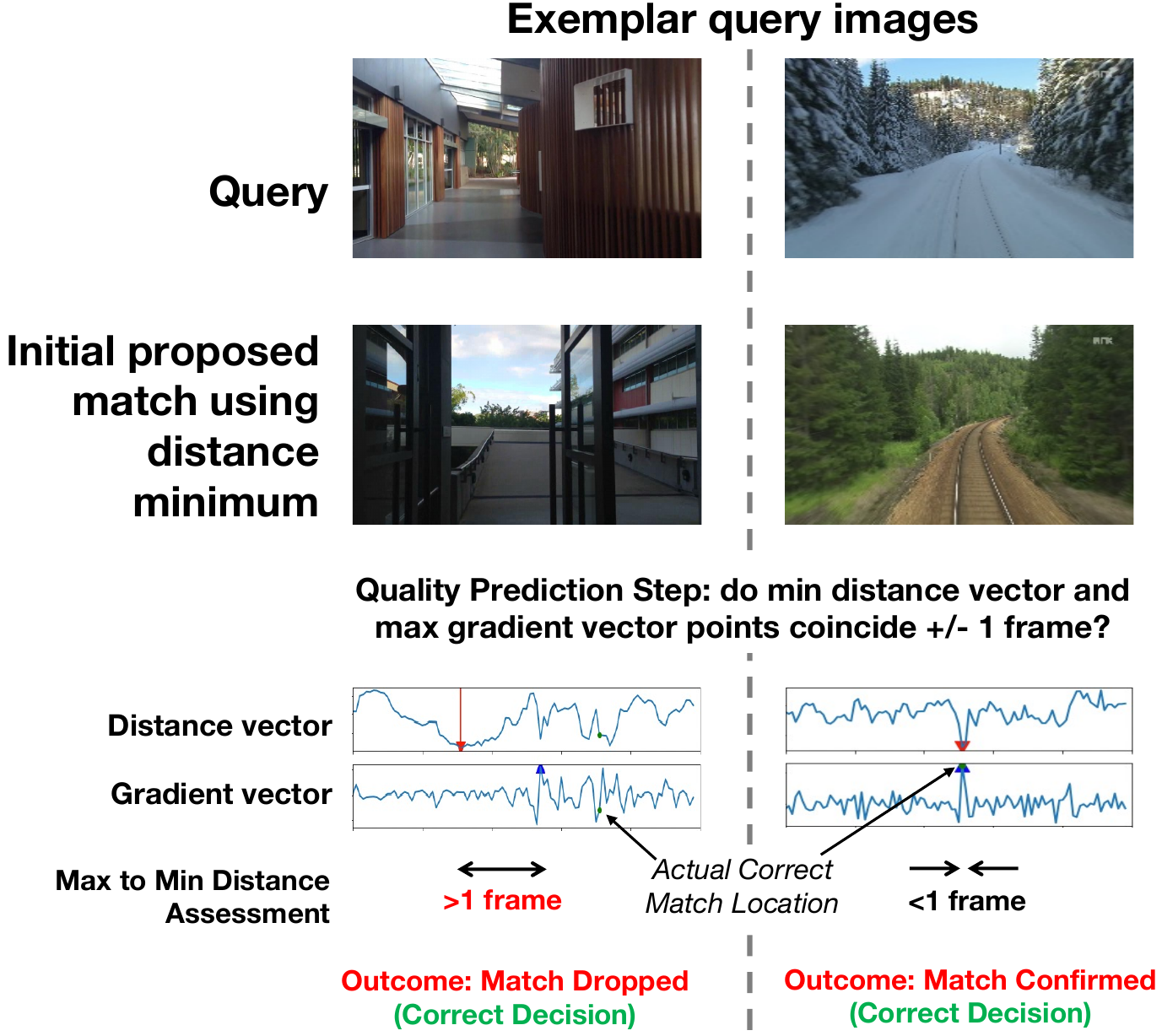}
\caption{Examples showing our prediction method applied to proposed ``good'' matches from a VPR technique. Consensus between the distance vector minimum (used by the VPR technique to generate the match) and gradient vector maximum is shown to predict a good quality (in-tolerance) match.}
\label{fig:pictures}
\end{figure}

\subsubsection{Formulation of Distance Matrix} 

For completeness sake, here we briefly introduce our distance matrix calculation, formed by computing the euclidean or cosine distance between the features in the current scene (query image), to features in a set of reference images. The candidate best match is then selected as the reference frame with the minimum distance to each query along a route, representing the best estimated location. In this paper we use a distance matrix \(\mathbf{D} \in \mathbb{R}^{n \times m}\) as a representation of feature euclidean distance between all \textit{m} queries and \textit{n} references, and a distance vector \(\mathbf{d}_{j}=\mathbf{D}[:,j]\) as the distance between a single query image and all reference images, which is the method used in real-time implementation.

Each element \(\mathbf{D}_{ij}\) is the distance between the features in the \(i^{th}\) database image and the \(j^{th}\) query image, and the reference with the minimum distance, \(d_0\), is found as the candidate best match where, \(i_{d0}= {\text{argmin}}(\mathbf{d}_{j})\). Lower values at the minimum point indicate a stronger match. 

\subsubsection{Formulation of Gradient Matrix}

We introduce in this work a gradient matrix, which is formulated by computing a modified average gradient around each point along the distance vector, computed as:
\begin{equation} \label{eq:grad_factor}
    g=\left\{
        \begin{array}{ll}
            \frac{1}{2}(d_{i+1}+d_{i-1})-d_{i}, & 1<i<n\\
            d_{i+1}-d_{i}, & i=1\\
            d_{i-1}-d_{i}, & i=n\\
        \end{array}
    \right.
\end{equation}

We use convolution to enhance the signal using a 3 \(\times\) 3 kernel, and pad the first two frames with the mean of the gradient vector. A potential negative impact of this in a real-time implementation may be decreased prediction accuracy for the first and second query frames.

A gradient vector, \(\mathbf{g}_{j}=\mathbf{G}[:,j]\) is formed for the \textit{j}th query, representing the rate of change between the query features, and features in each reference frame along the route. Note that the order of the reference frames must be preserved, but the query images can be encountered in any order in real-time. The maximum gradient for each query can then be hypothesised as an alternative best match candidate:
\begin{equation} \label{eq:gradient_best_match}
    i_{g0}=\text{argmax}(\mathbf{g}_{j})
\end{equation}

\subsubsection{Rejection of out-of-tolerance points}\label{sec:reject_oot_points}

To finalise our technique, we compare the best match generated by the minimum distance vector and maximum gradient vector, and reject matches where the number of frames between these are greater than one. We form a binary prediction vector \(\mathbf{y}_{pred} \in \{0,1\}^{m}\) from this comparison, where \(y_{pred}=1\) representation predicts the localization estimate is in-tolerance (True Positive) and \(y_{pred}=0\) represents an out-of-tolerance prediction (False Positive):
\begin{equation} \label{eq:y_pred}
    y_{pred}=
        \begin{cases}
            1, & |i_{g0} - i_{d0}| <= 1\\
            0, & \text{otherwise.}
        \end{cases}
\end{equation}

The prediction vector can be used as a mask to retain only predicted good candidate matches from the distance matrix. Examples of this consensus based prediction approach are shown in Figure \ref{fig:pictures}.

\subsection{Anchored/Weighted sequence}

The methods described so far have the potential to improve single-frame-based matching, and could be further enhanced by naive application of a sequence-based matching technique. However, a key new opportunity here we pursue is to weight the sequence matching process based on the predictions around the quality of the localization estimates. Here we incorporate our method into SeqSLAM \cite{SeqSLAM} by heavily weighting predicted good points prior to performing sequence matching. This artificially gives each sequence template containing predicted good points a much lower value than the standard implementation, which anchors the sequence around our prediction.

For each query, we weight each minimum value, \(d_0\), in the distance vector corresponding to a predicted good point by a weighting factor, \(w\), between 0 and 1 to create a weighted distance matrix, \(\mathbf{D}_w\):
\begin{equation} \label{eq:weighted_D}
    D_{w}\big|_{i=i_{d0},j}=
        \begin{cases}
            d_0 - w(d_0-D_{min}), & y_{pred}=1\\
            D_{i,j}, & y_{pred}=0\\
        \end{cases}
\end{equation}

The weighting factor \(w\) lowers the minimum value in the distance vector towards the overall minimum of the distance matrix. All other values, including predicted bad points (\(y_{pred}=0\)), are left unchanged. 

To implement sequence matching, we use the computationally efficient method of convolving the weighted distance matrix with an identity matrix with sequence length \(\textit{L}\), as described in \cite{garg2021seqmatchnet} and \cite{Fischer22}:
\begin{equation} \label{eq:convolution}
    \mathbf{D}_{\text{\textit{wseq}}}=\mathbf{I}_L \ast \mathbf{D}_{w},
\end{equation}
and identify the new best candidate sequence for each query: 
\begin{equation} \label{eq:new_gradient_best_match}
    i_{0}=\text{argmin}(\mathbf{D}_{\text{\textit{wseq}}}[:,j])
\end{equation}

Using a weighting factor \(w=1\) reduces all predicted `good' values to the minimum value in the distance matrix, \(\mathbf{D}_{min}\). To enable generation of PR-curves, a weighting factor \textit{w} less than one (e.g. \(w=0.99\)) is recommended. 

\section{Experimental Setup}\label{sec:method}

To test our hypotheses, we implement our approach using 15 combinations of datasets and VPR techniques, described in the sections below, and compare the performance of our method against those of baseline techniques. 

\subsection{VPR Techniques}

We apply our method to Sum-of-Absolute Differences (SAD), using patch-normalized downsampled images \cite{SeqSLAM}; NetVLAD \cite{NetVLAD18} based on a CNN architecture; and recent state-of-the-art approach PatchNetVLAD, which uses fusion of locally-global descriptors on multiple-scales using pre-trained models \cite{PatchNetVLAD}.

\subsection{Datasets}

Four widely used benchmark datasets have been used to implement our technique: Gardens Point Walking \cite{Chen14}, Nordland \cite{Olid18}, 4Seasons OfficeLoop \cite{wenzel2020fourseasons} and Oxford RobotCar \cite{RobotCarDatasetIJRR,RobotCarGroundTruth}. In each case, we use one reference set and one query set taken on the same route for our unsupervised prediction. When benchmarking against our previous supervised prediction technique, we use a second calibration reference/query set over a different route as described in \cite{Carson22}, along with ground-truth data for the calibration set.

Gardens Point Walking is a walking dataset obtained on a university campus with viewpoint variation between the reference and query sets. We use the first 100 frames with the left set as reference and right set as query, with 2.5m between each frame. Nordland contains images from a Norwegian railway, and uses a fixed viewpoint, but with large seasonal variation between reference and query sets. We have used two separate sets from this dataset: fall/spring and summer/winter as query/reference sets. The summer/winter set provides a challenging dataset due to significant differences in daylight levels, snow cover, and foliage along the route. In both cases we use 1150 images over 35km from Test Section~2 as defined in \cite{Olid18}, with mean distance of 30m between consecutive images, and tunnels and station stops removed.

4Seasons and Oxford RobotCar contain images recorded from passenger vehicles driving in urban environments. Challenges in these sets include presence of dynamic objects, viewpoint variations along the route (e.g. when changing lanes) and environmental differences. In 4Seasons, we use 381 frames for reference and query routes over 500m with little variation apart from dynamic objects on the road, providing an `easy' set that all chosen VPR techniques work well on. For Oxford RobotCar, we use 1800 image frames over approximately 2.25km recorded in sunny conditions as the reference set, in rain for the query set, with 1.25m median distance between frames \cite{Molloy20}. 

Together, the datasets chosen provide a broad set of conditions and challenges to assess the performance of our method.

\section{Results}\label{sec:results}

We present performance results for the overall combination of our two proposed contributions in Section \ref{sec:combined_results}, and the results of our two contributions independently in Sections \ref{sec:results_unsuper} and \ref{sec:results_weightedseq}. We also present an ablation study showing the effect of sequence length and weighting factor on the weighted sequence performance, and experimental results for mean execution time at inference.

\subsection{Combined system}\label{sec:combined_results}

Precision and recall are commonly used performance measures defined as ratios of in-tolerance localization matches (True Positives), out-of-tolerance matches (False Positives), and False Negatives (FN) (good matches that have been discarded), such that precision=\(\frac{TP}{TP+FP}\) and recall=\(\frac{TP}{TP+FN}\) \cite{Lowry2016}. Area Under the PR-curve (AUC) is a proxy for average precision, with higher values corresponding to higher performance. We impose a tight tolerance of +/-1 image frame as a ``good'' (in-tolerance) match in all experiments.

Table \ref{table:ap_20R_bl_seq_comparison} includes AUC up to 20\% recall for baseline sequences, and sequences weighted with our unsupervised predictions. In 13 out of 15 combinations our weighted sequence has higher or equal AUC than the baseline. A common weighting factor of 0.99 and sequence length of 2 is used in all cases.

\begin{figure*}
\centering
\includegraphics[width=7in]{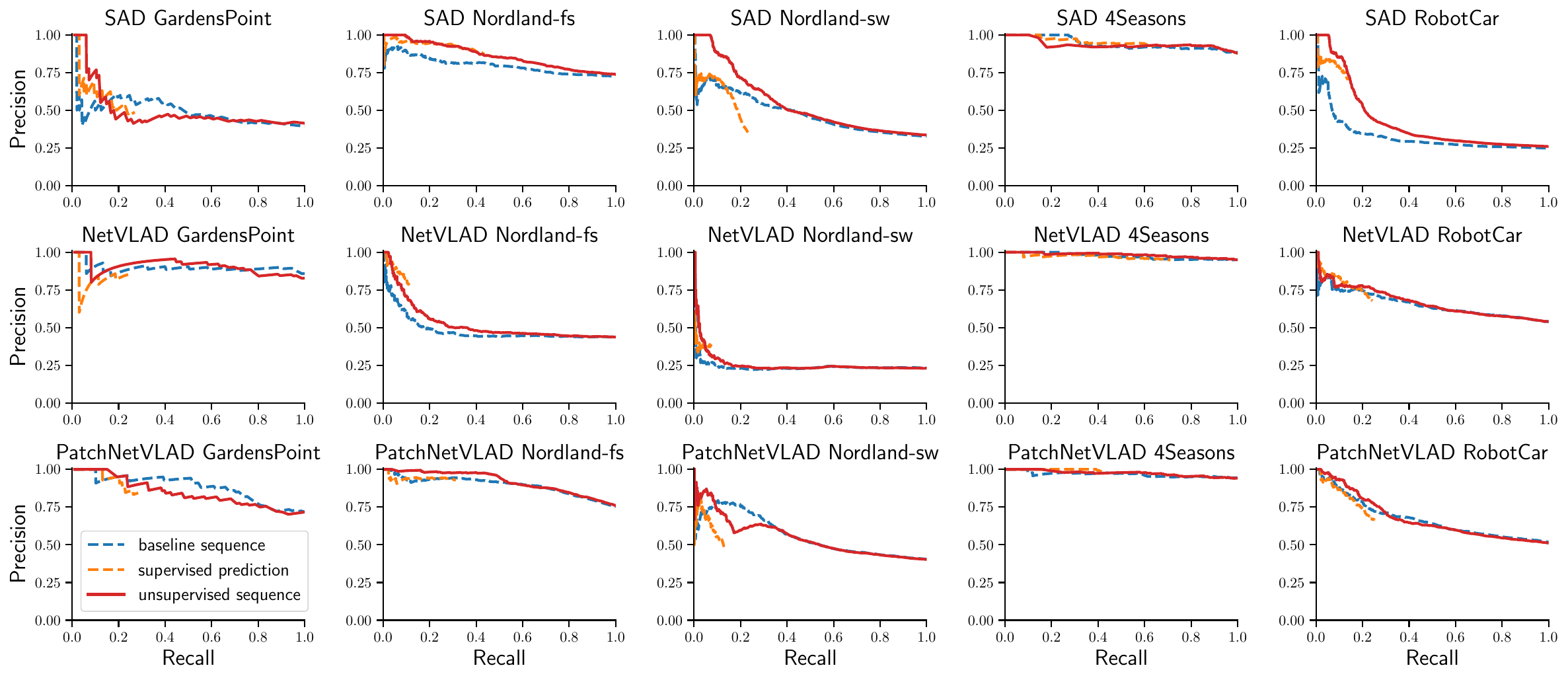}
\caption{PR-curves showing that in most cases, the performance impact of our proposed unsupervised approach using weighted sequences (combined system) is higher at high precision/low recall points compared to using a baseline (unweighted) sequence, over 15 combinations of varying datasets and VPR techniques. We also show comparison to the supervised prediction technique (without sequencing), from our previous work.}
\label{fig:pr_curves_overall}
\end{figure*}

PR-curves for all three VPR techniques applied to each dataset are shown in Figure \ref{fig:pr_curves_overall} for the baseline sequences using no weighting, weighted sequences using the unsupervised prediction method, and the supervised method from our previous work. In the majority of cases, the PR-curve using weighted sequences has higher precision for the same recall as the unweighted sequences in the higher precision range for each dataset, which is the operating point of interest for safety-critical scenarios. The performance using weighted sequences also matches or exceeds the precision of our supervised technique in several cases, albeit up to a lower recall. 

\begin{table}[!ht]
\renewcommand{\arraystretch}{1.2}
\caption{Comparison of AUC up to 20\% Recall for baseline (BL) and weighted unsupervised 
sequences (Ours) with sequence length 2 (higher is better).} 
\label{table:ap_20R_bl_seq_comparison}
\centering
\begin{tabular}{|l|c c|c c|c c|}
\multicolumn{1}{c}{} & \multicolumn{2}{c}{SAD} & \multicolumn{2}{c}{NetVLAD} & \multicolumn{2}{c}{PatchNetVLAD}\\
\hline
Dataset & BL & Ours & BL & Ours & BL & Ours\\
\hline
GardensPoint & 0.54             & \textbf{0.68}     & \textbf{0.87} & \textbf{0.87}     & 0.92              & \textbf{0.94}\\
Nordland-fs  & 0.88             & \textbf{0.98}     & 0.62          & \textbf{0.74}     & 0.95              & \textbf{0.99}\\
Nordland-sw  & 0.66             & \textbf{0.89}     & 0.27          & \textbf{0.35}     & \textbf{0.74}     & \textbf{0.74}\\
4seasons     & \textbf{0.99}    & 0.96              & \textbf{0.99} & 0.98              & 0.97              & \textbf{0.98}\\
RobotCar     & 0.50             & \textbf{0.82}     & 0.78          & \textbf{0.81}     & 0.88              & \textbf{0.92}\\
\hline
Average & \multicolumn{3}{|c}{Baseline 0.77} & \multicolumn{3}{c|}{Ours \textbf{0.84}}\\
\hline
\end{tabular}
\end{table}

\subsection{Performance of individual contributions}

The following sections present an analysis of results for our two contributions independently.

\subsubsection{Unsupervised Prediction}\label{sec:results_unsuper}

We compare our unsupervised method to the following: (1) baseline VPR technique, sweeping the minimum distance as a threshold, and (2) applying a supervised (rather than unsupervised) prediction to the baseline VPR technique, and (3) generating matches using the maximum value in the gradient matrix only, and sweeping the maximum gradient as a threshold. 

\begin{figure}[!t]
\centerline{\includegraphics[width=3in]{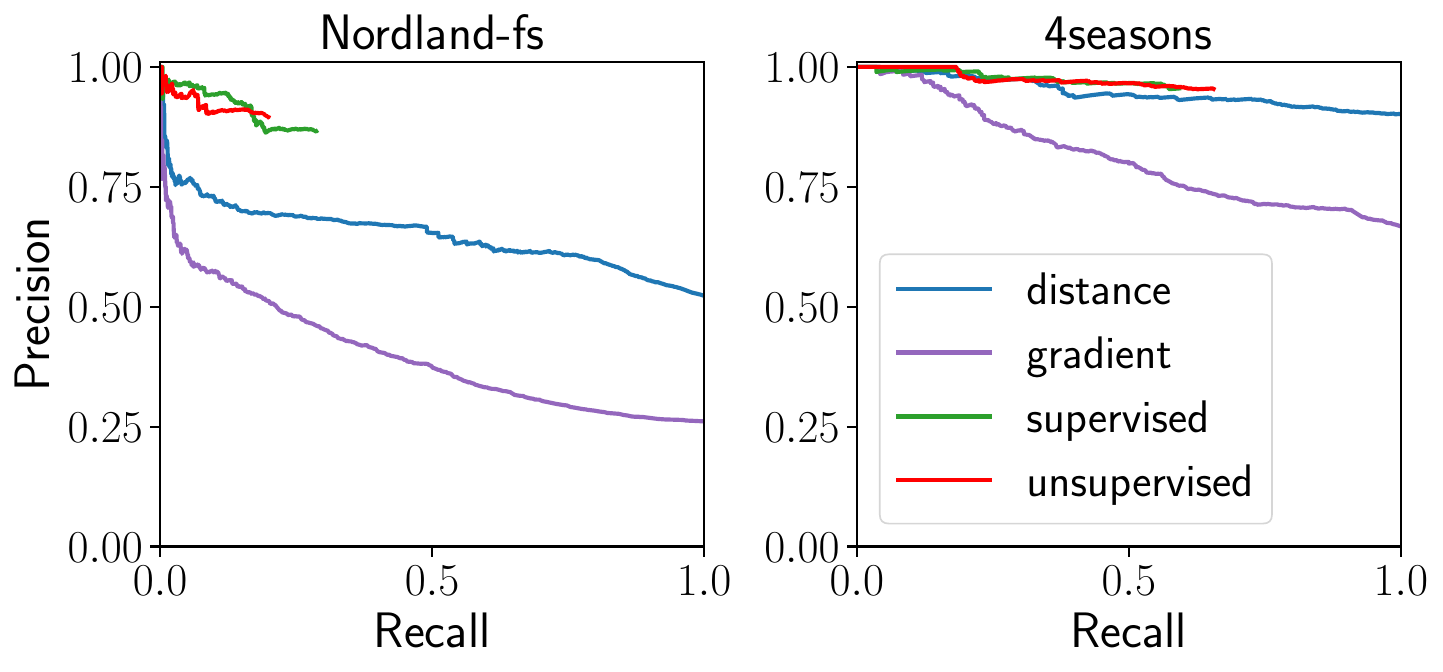}}
\caption{PR-curves for all three VPR techniques averaged over the Nordland fall/spring and 4Seasons datasets, showing the superior performance using the minimum distance (baseline, blue) versus the maximum gradient only (purple) to determine best matches, and also the improved precision and impact on recall by retaining only `good' matches using both supervised prediction (green) and our unsupervised prediction (red).}
\label{fig:pr_curves_predictions}
\end{figure}

Fig. \ref{fig:pr_curves_predictions} demonstrates the impact of these comparisons as  averaged PR-curves for the Nordland fall/spring and 4Seasons datasets, and shows the increased precision over the baseline when supervised and unsupervised predictions are used to remove `bad' matches. The curves are truncated to a maximum recall value due to many matches being discarded, such that 100\% recall cannot be achieved. 

These results also show that for these two datasets, using the maximum gradient alone to determine the best match produces decreased performance compared to the baseline, which uses the minimum distance value. Increased precision is only achieved when the match with the minimum distance and maximum gradient coincide (within +/1 one frame), which is the basis for our unsupervised technique.

\subsubsection{Weighted Sequences}\label{sec:results_weightedseq}

The performance of the weighted sequence is directly impacted by the prediction quality, chosen weighting factor, and is also bounded in part by the performance of the underlying VPR technique in the environment. 
A break-even point will exist for each combination of these characteristics such that below a particular prediction quality threshold, use of the weighted sequence may actually \textit{decrease} the combined system performance. The weighting factor, \(w\), tempers the impact of the prediction on performance, and can be chosen based on the estimated performance quality of a given prediction method and dataset / VPR technique combination. 

\begin{figure}
\centerline{\includegraphics[width=2.5in]{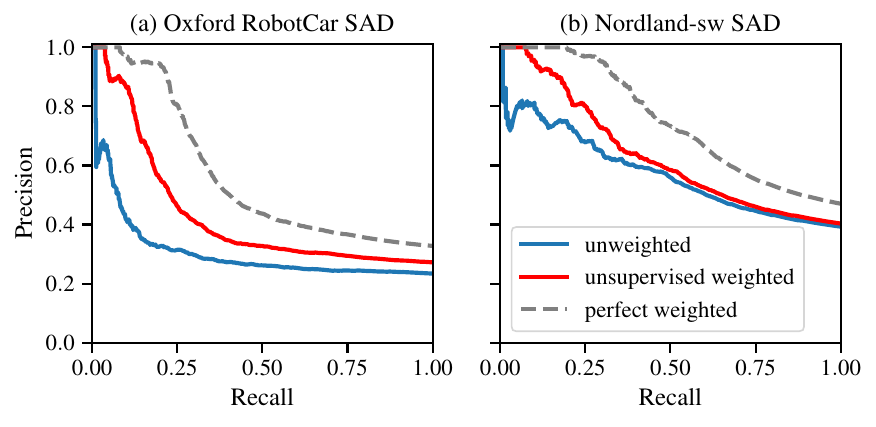}}
\caption{PR-curves showing the performance improvement of weighted sequences (using unsupervised predictions) over unweighted sequences (baseline), and the potential combined system performance (dashed) that could be obtained with perfect prediction on this specific route (with sequence length = 3) is shown for comparison.}
\label{fig:weighted_sequences}
\end{figure}

Figure \ref{fig:weighted_sequences} shows the increase in performance by using a weighted sequence, where the unsupervised method (from Section~\ref{sec:reject_oot_points}) is used to weight the distance matrix at predicted `good' locations, for two datasets using SAD. The potential increase on this specific route and conditions using a perfect prediction is also shown to illustrate how, even with unrealistic perfect prediction, upper bound performance is limited by the underlying system. 

To highlight the impact of this precision increase on the number of false positive matches, we can examine a specific example from our results. In Fig. \ref{fig:weighted_sequences}(b), at 20\% recall, the number of false positives (incorrect matches) in the GardensPoint SAD set is 37 for the unweighted sequence but drops to only 17 for the weighted sequence using unsupervised predictions.

\begin{figure}
\centerline{\includegraphics[width=3in]{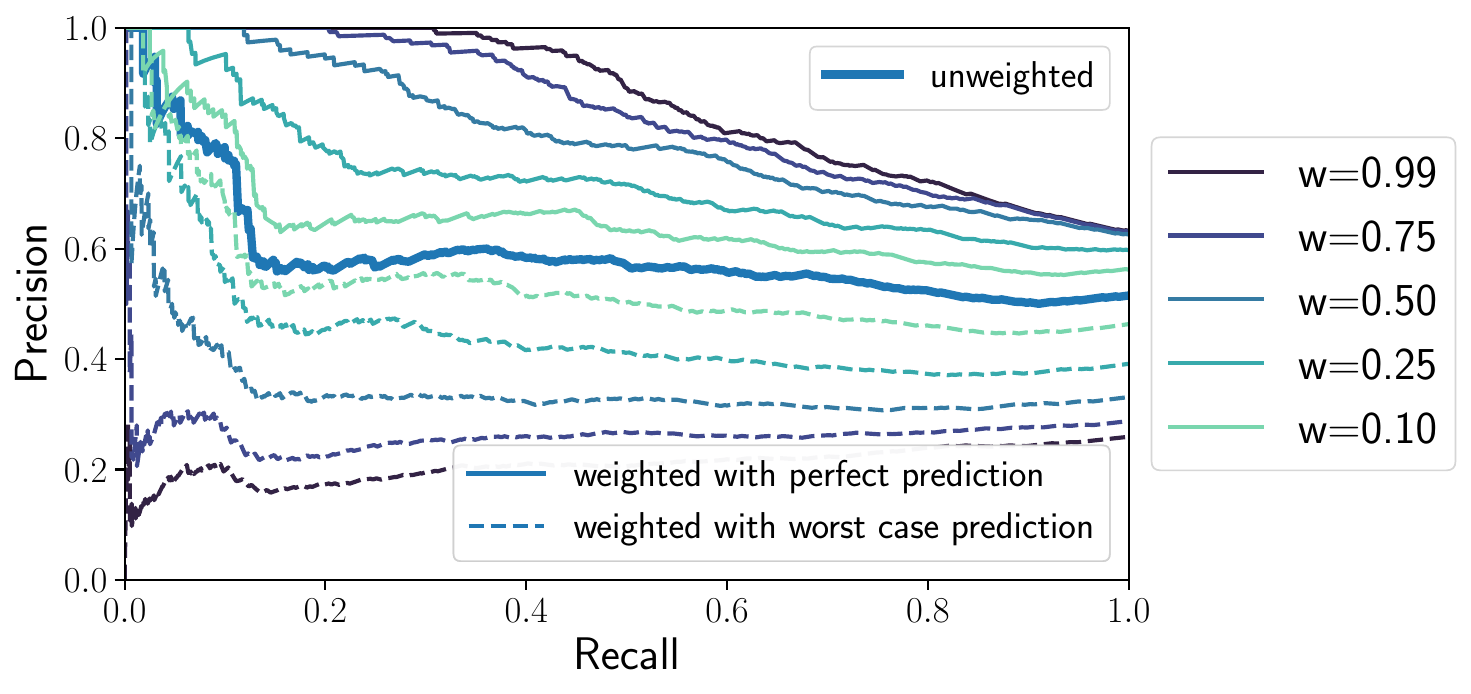}}
\caption{PR-curves characterizing the relationship between performance prediction and weighting factor for one example route in the Oxford RobotCar dataset using SAD, for both a high quality and poor quality prediction system. Above the unweighted thicker middle line, we have a high quality predictor: as weighted use of the predictions goes up, performance goes up. Below the unweighted line, we have a poor quality predictor: as weighting goes up, performance actually decreases. }
\label{fig:varying_w}
\end{figure}

Finally, Figure \ref{fig:varying_w} presents a characterisation of the impact of the prediction performance over a range of weighting factors, for two different scenarios, a good prediction system and a poor prediction system, using the Oxford Robotcar dataset. The thicker line in the middle of the graph shows an unweighted sequence matcher as a baseline. \textit{Below} the unweighted line, we have a badly performing predictor, showing that as the weighting factor goes up, the performance gets progressively worse. Above the unweighted line, we have a good predictor; we can see as the weighting factor goes up in this case, the overall performance goes up. 

\subsubsection{Ablation Study}

\begin{figure}[!t]
\centerline{\includegraphics[width=3in]{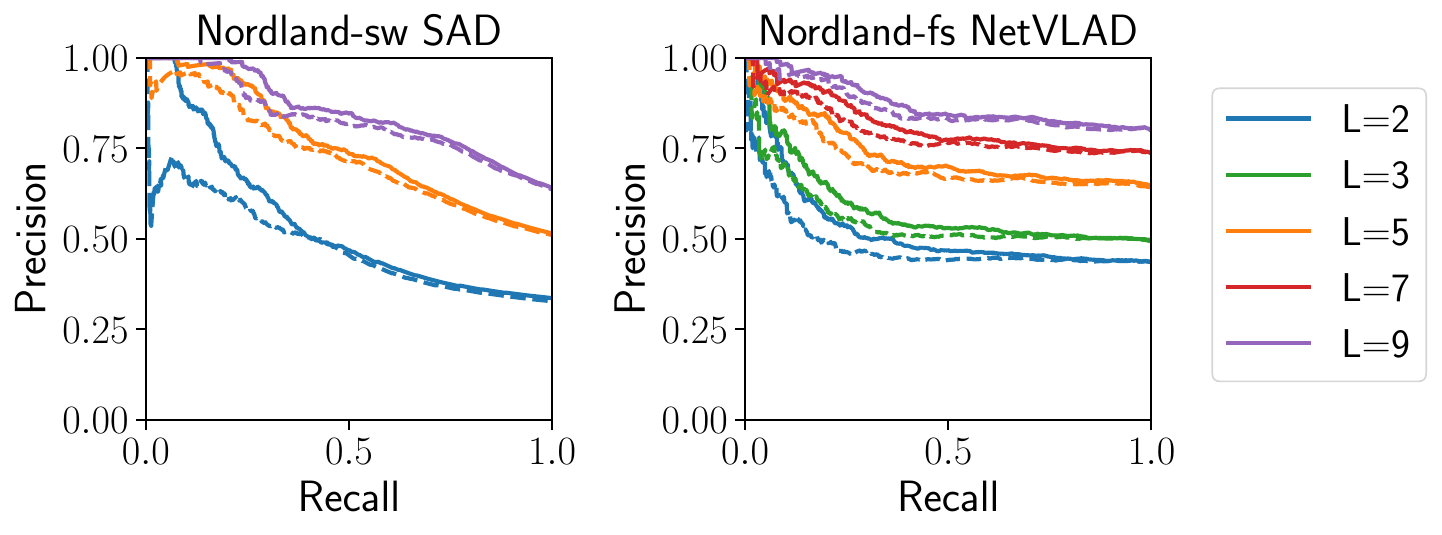}}
\caption{PR-curves for Nordland-fall/spring dataset with NetVLAD, showing the improved performance for  weighted sequences (solid) vs baseline (unweighted) sequences (dashed) for various sequence lengths, with constant weighting factor.}
\label{fig:ablation}
\end{figure}

An ablation study of weighted sequences for varying sequence lengths shows that the performance increase persists as sequence length increases, but the relative improvement over the baseline decreases. Examples are given in Figure \ref{fig:ablation} for two combinations of Nordland datasets with varying sequence lengths from L=2 to L=9.

\subsection{Computational resources}\label{sec:computational_resources}

To evaluate whether our methods are suitable for real-time implementation, we analyse scalability and present experimental results showing mean execution time for our method.

\subsubsection{Theoretical explanation of scalability} 

At run-time, our prediction method requires generation of a gradient vector for each query point. Both the prediction and weighted sequence filter require convolution to be performed along the gradient and distance vectors. Execution time will therefore increase linearly with reference database size, as the same arithmetic operations are performed for each reference image.

\subsubsection{Execution time at inference} 

\begin{figure}
    \centering
    \includegraphics[width=2.25in]{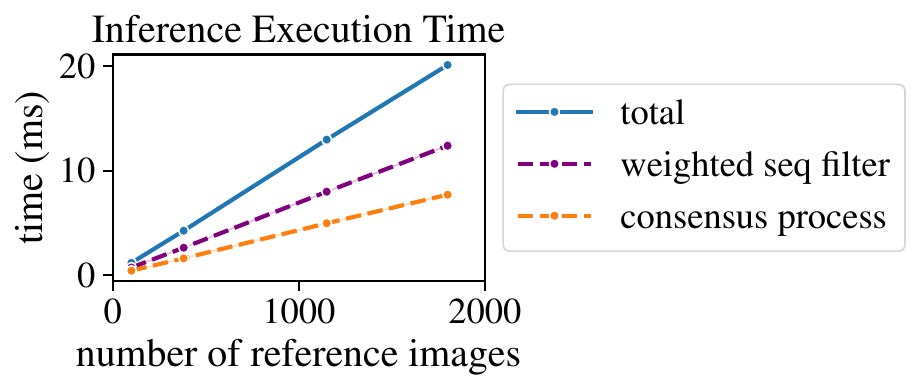}
    \caption{Mean execution time at inference showing the separate contributions of our two proposed methods, suggesting the method is suitable for real-time applications (under 20ms for up to 1800 reference frames, L=3).}
    \label{fig:execution_time}
\end{figure}

Mean execution time for real-time implementation in Python has been measured running on an Apple Silicon M1 chip with 16GB RAM. Figure \ref{fig:execution_time} shows the total mean execution time required for the unsupervised prediction and weighted sequence at inference is less than 20ms per query for datasets presented here up to 1800 reference frames. This includes the time required to implement the sequence filter, which is common with the baseline sequence technique. Note that our implementation is not optimised and can be considered a worst case. Results in Fig~\ref{fig:execution_time} provide evidence that our proposed methods are suitable for implementation in real-time applications.

\section{Discussion}\label{sec:discussion}

We have presented two complementary methods for improving VPR performance based on: (1) unsupervised prediction of `good' localization estimates along a route, and (2) using sequences weighted with predictions. Both methods improve performance but synergize in particular because the predictions enable the weighted nature of the modified sequence matching technique. Over a comprehensive combination of benchmark datasets and VPR techniques, results show that both techniques lead to improvements in localization performance. A limitation is that these techniques are best suited to applications where high precision, low recall operating points are acceptable.

There are a number of exciting avenues for future work. While the unsupervised nature of the approach presented here has practical advantages, there are a subset of applications where supervised training of a prediction system is feasible. In such situations, it would be interesting to examine how fusing both supervised and unsupervised methods could further improve both performance and versatility. In particular, an unsupervised method could potentially provide an online training signal for a supervised method. In addition, while difference and gradient-based matching and prediction has been shown to be effective here, it may be possible to learn more abstract ``factors" that provide further predictive capability around localization performance. Finally, while we have used prediction outputs to weight the sequence matching process, it may be possible to more organically link the prediction processes and weighted sequential matching processes into a single monolithic system, with further potential advantages.

\bibliographystyle{IEEEtran}
\bibliography{IEEEabrv,ref.bib}

\end{document}